\def\year{2021}\relax
\documentclass[letterpaper]{article} % DO NOT CHANGE THIS
\usepackage{aaai21}  % DO NOT CHANGE THIS
\usepackage{times}  % DO NOT CHANGE THIS
\usepackage{helvet} % DO NOT CHANGE THIS
\usepackage{courier}  % DO NOT CHANGE THIS
\usepackage[hyphens]{url}  % DO NOT CHANGE THIS
\usepackage{graphicx} % DO NOT CHANGE THIS
\usepackage{natbib}  % DO NOT CHANGE THIS AND DO NOT ADD ANY OPTIONS TO IT
\usepackage{caption} % DO NOT CHANGE THIS AND DO NOT ADD ANY OPTIONS TO IT
\usepackage{booktabs}       % professional-quality tables
\usepackage{amsfonts}       % blackboard math symbols
\usepackage{microtype}      % microtypography
\usepackage[dvipsnames]{xcolor}
\usepackage{url}
\usepackage{verbatim} % allows multiline comments
\usepackage{graphicx}
\usepackage{multirow}
\usepackage{algorithm}
\usepackage[noend]{algorithmic}
\usepackage{xspace}
\usepackage{epsfig}
\usepackage{amsmath}
\usepackage{amsthm}
\usepackage{amssymb}
\usepackage{times}
\usepackage{xr}
\usepackage{bbm}
\usepackage{bm}
\usepackage{subcaption}
\usepackage{enumitem}
\usepackage{hyperref}       % hyperlinks
\usepackage{multicol}
\usepackage{tabularx}
\usepackage{perpage} %the perpage package
\MakePerPage{footnote} %the perpage package command
\usepackage{wrapfig}

\newcommand{\xhdr}[1]{{\noindent\bfseries #1}.}

\newcommand{\mb}{\mathbf}
\newcommand{\cut}[1]{}

\newtheorem{proposition}{Proposition}

\newcommand{\ie}{\emph{i.e.}}

\usepackage{amsmath,amsfonts,bm}

\def\eqref#1{equation~\ref{#1}}
\def\1{\bm{1}}

\DeclareMathAlphabet{\mathsfit}{\encodingdefault}{\sfdefault}{m}{sl}
\SetMathAlphabet{\mathsfit}{bold}{\encodingdefault}{\sfdefault}{bx}{n}

\title{Identity-aware Graph Neural Networks}
\author{
    Jiaxuan You, Jonathan Gomes-Selman, Rex Ying, Jure Leskovec 
    \\
}
\affiliations{
    Department of Computer Science, Stanford University \\
    \{jiaxuan, jgs8, rexy, jure\}@cs.stanford.edu
}
\begin{document}

\maketitle

\begin{abstract}

Message passing Graph Neural Networks (GNNs) provide a powerful modeling framework for relational data. However, the expressive power of existing GNNs is upper-bounded by the 1-Weisfeiler-Lehman (1-WL) graph isomorphism test, which means GNNs that are not able to predict node clustering coefficients and shortest path distances, and cannot differentiate between different $d$-regular graphs.
Here we develop a class of message passing GNNs, named \emph{Identity-aware Graph Neural Networks (ID-GNNs)}, with greater expressive power than the 1-WL test. ID-GNN offers a minimal but powerful solution to limitations of existing GNNs.
ID-GNN extends existing GNN architectures by inductively considering nodes' identities during message passing.
To embed a given node, ID-GNN first extracts the ego network centered at the node, then conducts rounds of \emph{heterogeneous message passing}, 
where different sets of parameters are applied to the center node than to other surrounding nodes in the ego network. 
We further propose a simplified but faster version of ID-GNN that injects node identity information as augmented node features.
Altogether, both versions of ID-GNN represent general extensions of message passing GNNs, where experiments show that transforming existing GNNs to ID-GNNs yields on average 40\% accuracy improvement on challenging node, edge, and graph property prediction tasks; 3\% accuracy improvement on node and graph classification benchmarks; and 15\% ROC AUC improvement on real-world link prediction tasks. Additionally, ID-GNNs demonstrate improved or comparable performance over other task-specific graph networks.
\end{abstract}

\section{Introduction}
Graph Neural Networks (GNNs) represent a powerful learning paradigm that have achieved great success \cite{scarselli2008graph,li2015gated,kipf2016semi,hamilton2017inductive,velickovic2017graph,xu2018powerful,you2020design}. Among these models, messaging passing GNNs, such as GCN \cite{kipf2016semi}, GraphSAGE \cite{hamilton2017inductive}, and GAT \cite{velickovic2017graph}, are dominantly used today due to their simplicity, efficiency and strong performance in real-world applications \cite{zitnik2017predicting,ying2018graph,you2018graph,you2019g2sat,you2020graph,you2020handling}.
The central idea behind message passing GNNs is to learn node embeddings via the repeated aggregation of information from local node neighborhoods using non-linear transformations \cite{battaglia2018relational}. %In this paper, we follow the practice of using the shorter term ``GNNs'' to refer to message passing GNNs \cite{xu2018powerful}, whenever possible.

\begin{figure*}[t]
\centering
\includegraphics[width=0.95\linewidth]{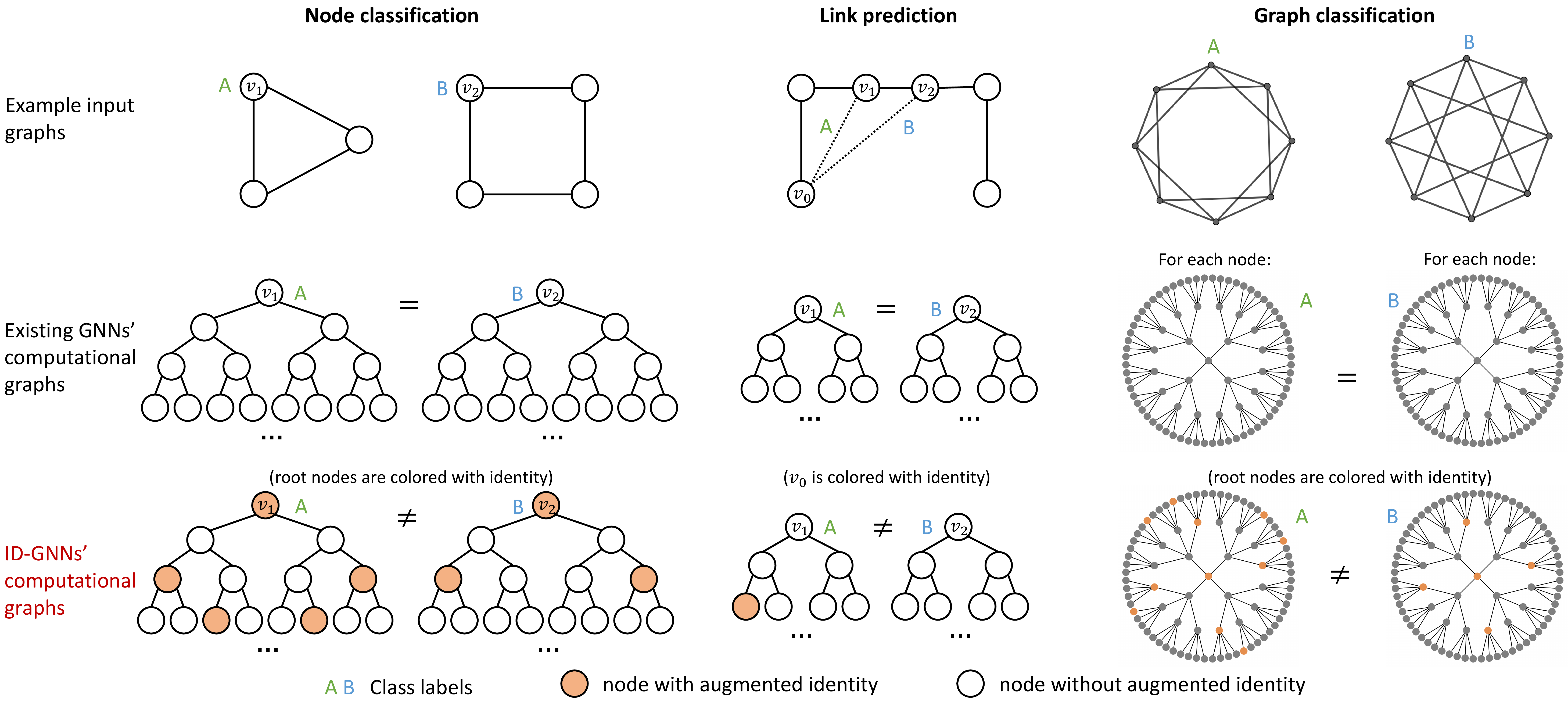} 
\vspace{-2mm}
\caption{
An overview of the proposed ID-GNN model. 
We consider node, edge and graph level tasks, and assume nodes do not have discriminative features.
Across all examples, the task requires an embedding that allows for the differentiation of nodes labeled $A$ vs. $B$ in their respective graphs. However, across all tasks, existing GNNs, regardless of their depth, will \textit{always} assign the same embedding to both nodes $A$ and $B$, because for all tasks the computational graphs are identical (middle row). In contrast, the colored computational graphs provided by ID-GNN allow for clear differentiation between the nodes of label $A$ and label $B$, as the colored computational graphs are no longer identical across the tasks.  
}
\label{fig:overview}
\end{figure*}

Although GNNs represent a powerful learning paradigm, it has been shown that the expressive power of existing GNNs is upper-bounded by the 1-Weisfeiler-Lehman (1-WL) test \cite{xu2018powerful}.
Concretely, a fundamental limitation of existing GNNs is that two nodes with different neighborhood structure can have the same computational graph, thus appearing indistinguishable. Here, a computational graph specifies the procedure to produce a node's embedding. Such failure cases are abundant (Figure \ref{fig:overview}): in node classification tasks, existing GNNs fail to distinguish nodes that reside in $d$-regular graphs of different sizes; in link prediction tasks, they cannot differentiate node candidates with the same neighborhood structures but different shortest path distance to the source node; and in graph classification tasks, they cannot differentiate $d$-regular graphs \cite{chen2019equivalence,murphy2019relational}. 
While task-specific feature augmentation can be used to mitigate these failure modes, the process of discovering meaningful features for different tasks is not generic and can, for example, hamper the inductive power of GNNs.

Several recent methods aim to overcome these limitations in existing GNNs. For graph classification tasks, a collection of works propose novel architectures more expressive than the 1-WL test \cite{chen2019equivalence,maron2019provably,murphy2019relational}. For link level tasks, P-GNNs are proposed to overcome the limitation of existing GNNs \cite{you2019position}. While these methods have a rich theoretical grounding, they are often task specific (either graph or link level) and often suffer from increased complexity in computation or implementation. 
In contrast, message passing GNNs have a track record of high predictive performance across node, link, and graph level tasks, while being simple and efficient to implement.
Therefore, extending message passing GNNs beyond the expressiveness of 1-WL test, to overcome current GNN limitations, is a problem of high importance. 

% \jure{Let's always use singular not plural. So, ID-GNN and not ID-GNNs. Because we are talking about a particular ID-GNN and not an entire class of models (like GNNs are a class of models)).}

\xhdr{Present work} Here we propose Identity-aware Graph Neural Networks (ID-GNNs), a class of \emph{message passing GNNs with expressive power beyond the 1-WL test}\footnote{Project website with code: \url{http://snap.stanford.edu/idgnn}}.
ID-GNN provides a universal extension and makes \emph{any} existing message passing GNN more expressive.
ID-GNN embeds each node by \emph{inductively} taking into account its identity during message passing. The approach is different from labeling each node with a one-hot encoding, which is \emph{transductive} (cannot generalize to unseen graphs).
As shown in Figure \ref{fig:overview}, we use an \emph{inductive identity coloring} technique to distinguish a node itself (the root node in the computational graph) from other nodes in its local neighborhood, within its respective computational graph. 
This added identity information allows ID-GNN to distinguish what would be identical computational graphs across node, edge and graph level tasks, and this way overcome the previously discussed limitations.

We propose two versions of ID-GNN.
As a general approach, identity information is incorporated by applying rounds of \emph{heterogeneous message passing}. Specifically, to embed a given node, ID-GNN first extracts the ego network centered at that node, then applies message passing, where the messages from the center node (colored nodes in Figure \ref{fig:overview}) and the rest of the nodes are computed using \emph{different sets of parameters}. This approach naturally applies to applications involving node or edge features. 
We also consider a simplified version of ID-GNN, where we inject identity information via cycle counts originating from a given node as augmented node features. These cycle counts capture node identity information by counting the colored nodes within each layer of the ID-GNN computational graph, and can be efficiently computed by powers of a graph's adjacency matrix. 

We compare ID-GNNs against GNNs across 8 datasets and 6 different tasks.
First, we consider a collection of challenging graph property prediction tasks where existing GNNs fail, including predicting node clustering coefficient, predicting shortest path distance, and differentiating random $d$-regular graphs. Then, we further apply ID-GNNs to real-world datasets.
Results show that transforming existing GNNs to their ID-GNN versions yields on average 40\% accuracy improvement on challenging node, edge, and graph property prediction tasks; 3\% accuracy improvement on node and graph classification benchmarks; and 15\% ROC AUC improvement on real-world link prediction tasks.
Additionally, we compare ID-GNNs against other expressive graph networks that are specifically designed for edge or graph-level tasks.
ID-GNNs demonstrate improved or comparable performance over these models, further emphasizing the versatility of ID-GNNs.

Our key contribution includes:
\textbf{(1)}
We show that message passing GNNs can have expressive power beyond 1-WL test.
\textbf{(2)}
We propose ID-GNNs as a general solution to the limitations in existing GNNs, with rich theoretical and experimental results.
\textbf{(3)}
We present synthetic and real world tasks to reveal the failure modes of existing GNNs and demonstrate the superior performance of ID-GNNs over both existing GNNs and other powerful graph networks.

% \jure{Cite Distance Encoding: Design Provably More Powerful Neural Networks for Graph Representation Learning.}
\section{Related Work}

\xhdr{Expressive neural networks beyond 1-WL test}
Recently, many neural networks have been proposed with expressive power beyond the 1-WL test, including \cite{chen2019equivalence, maron2019provably, murphy2019relational,you2019position,li2020distance}. However, these papers introduce extra, often task/domain specific, components beyond standard message passing GNNs. For example, P-GNN's embeddings are tied with random anchor-sets and, thus, are not applicable to node/graph level tasks which require deterministic node embeddings \cite{you2019position}. 
In this paper we emphasize the advantageous characteristics of message passing GNNs, and show that GNNs, after incorporating inductive identity information, can surpass the expressive power of the 1-WL test while maintaining benefits of efficiency, simplicity, and broad applicability.

\xhdr{Graph Neural Networks with inductive coloring}
Several models color nodes with augmented features to boost existing GNNs' performance \cite{xu2019can,velikovi2020neural,zhang2018link}.
However, existing coloring techniques are problem and domain-specific (\emph{i.e.} link prediction, algorithm execution), and are not generally applicable to node and graph-level tasks. In contrast,
ID-GNN is a general model that can be applied to any node, edge, and graph level task. It further adopts a heterogeneous message passing approach, which is fully compatible to cases where nodes or edges have rich features.

\xhdr{GNNs with anisotropic message passing}
We emphasize that ID-GNNs are fundamentally different from GNNs based on anisotropic message passing, where different attention weights are applied to different incoming edges \cite{bresson2017residual,hamilton2017inductive,monti2017geometric,velickovic2017graph}. 
Adding anisotropic message passing does not change the underlying computational graph because the same message passing function is symmetrically applied across all nodes.
Therefore, these models still exhibit the limitations summarized in Figure \ref{fig:overview}.
\section{Preliminaries}

A graph can be represented as $\mathcal{G} = (\mathcal{V},\mathcal{E})$, where $\mathcal{V} = \{v_1, ..., v_n\}$ is the node set and $\mathcal{E} \subseteq \mathcal{V} \times \mathcal{V}$ is the edge set. Nodes can be paired with features $\mathcal{X} = \{\mb{x}_v | \forall v \in \mathcal{V}\}$, and edges can have features $\mathcal{F} = \{\mb{f}_{uv} | \forall e_{uv} \in \mathcal{E}\}$.
As discussed earlier, we focus on message passing GNNs throughout this paper. We follow the definition of GNNs in \cite{xu2018powerful}.
The goal of a GNN is to learn meaningful node embeddings $\mb{h}_v$ based on an iterative aggregation of local network neighborhoods. The $k$-th iteration of message passing, or the $k$-th layer of a GNN, can be written as:
\begin{equation}
    \label{eq:gnn}
    \begin{aligned}
    & \mb{m}_u^{(k)} = \textsc{Msg}^{(k)}(\mb{h}_u^{(k-1)}), \\
    & \mb{h}_v^{(k)} = \textsc{Agg}^{(k)}\big(\{\mb{m}_u^{(k)}, u \in \mathcal{N}(v)\}, \mb{h}_v^{(k-1)}\big)
    \end{aligned}
\end{equation}
where $\mb{h}_v^{(k)}$ is the node embedding after $k$ iterations, $\mb{h}_v^{(0)}=\mb{x}_v$, $\mb{m}_v^{(k)}$ is the message embedding, and $\mathcal{N}(v)$ is the local neighborhood of $v$. Different GNNs have varied definitions of $\textsc{Msg}^{(k)}(\cdot)$ and $\textsc{Agg}^{(k)}(\cdot)$.
For example, a GraphSAGE uses the definition ($\mb{W}^{(k)}$, $\mb{U}^{(k)}$ are trainable weights):
\begin{align}
    & \mb{m}_u^{(k)} = \textsc{ReLU}(\mb{W}^{(k)}\mb{h}_u^{(k-1)}), \\
    & \mb{h}_v^{(k)} = \mb{U}^{(k)}\textsc{Concat}\big(\textsc{Max}\big(\{\mb{m}_u^{(k)},u \in \mathcal{N}(v)\}\big), \mb{h}_v^{(k-1)}\big) \notag
\end{align}
The node embeddings $\mb{h}_v^{(K)}, \forall v \in \mathcal{V}$ are then used for node, edge, and graph level prediction tasks.

\section{Identity-aware Graph Neural Networks}
\subsection{ID-GNNs: GNNs beyond the 1-WL test}
We design ID-GNN so that it can make \emph{any} message passing GNN more expressive. ID-GNN is built with two important components: \textbf{(1)} \emph{inductive identity coloring} where identity information is injected to each node, and \textbf{(2)} \emph{heterogeneous message passing} where the identity information is utilized in message passing. Algorithm 1 provides an overview.

\xhdr{Inductive identity coloring}
To embed a given node $v\in\mathcal{G}$ using a $K$-layer ID-GNN, we first extract the $K$-hop ego network $\mathcal{G}_v^{(K)}$ of $v$. We then assign a unique coloring to the central node of the ego network $\mathcal{G}_v^{(K)}$. 
Altogether, nodes in $\mathcal{G}_v^{(K)}$ can be categorized into two types throughout the embedding process: nodes with coloring and nodes without coloring.
This coloring technique is \emph{inductive} because even if nodes are permuted, the center node of the ego network can still be differentiated from other neighboring nodes.

\xhdr{Heterogeneous message passing}
$K$ rounds of message passing are then applied to all the extracted ego networks. To embed node $u \in \mathcal{G}_v^{(K)}$, we extend Eq. \ref{eq:gnn} to enable heterogeneous message passing:
\begin{equation}
    \label{eq:idgnn}
    \begin{aligned}
    & \mb{m}_s^{(k)} = \textsc{Msg}^{(k)}_{\mathbbm{1}[s = v]}(\mb{h}_s^{(k-1)}), \\
    & \mb{h}_u^{(k)} = \textsc{Agg}^{(k)}\big(\{\mb{m}_s^{(k)}, s \in \mathcal{N}(u)\}, \mb{h}_u^{(k-1)}\big)
    \end{aligned}
\end{equation}
where only $\mb{h}_v^{(K)}$ is used as the embedding representation for node $v$ after applying $K$ rounds of Eq. \ref{eq:idgnn}.
Different from Eq. \ref{eq:gnn}, two sets of $\textsc{Msg}^{(k)}$ functions are used, where $\textsc{Msg}^{(k)}_1(\cdot)$ is applied to nodes with identity coloring, and $\textsc{Msg}^{(k)}_0(\cdot)$ is used for node without coloring. The indicator function $\mathbbm{1}[s = v] = 1 \text{ if }s = v \text{ else } 0$ is used to index the selection of these functions. This way, the inductive identity coloring is \emph{encoded into the ID-GNN computational graph}.

A benefit of this heterogeneous message passing approach is that it is \emph{applicable to any message passing GNN}. For example, consider the following message passing scheme, which extends the definition of GNNs in Eq. \ref{eq:idgnn} by including edge attributes $\mb{f}_{su}$ during message passing:
\begin{equation}
    \begin{aligned}
    & \mb{m}_{su}^{(k)} = \textsc{Msg}^{(k)}_{\mathbbm{1}[s = v]}(\mb{h}_s^{(k-1)}, \mb{f}_{su}), \\
    & \mb{h}_u^{(k)} = \textsc{Agg}^{(k)}\big(\{\mb{m}_{su}^{(k)}, s \in \mathcal{N}(u)\}, \mb{h}_u^{(k-1)}\big)
    \end{aligned}
\end{equation}

\setlength{\textfloatsep}{4mm}% Remove \textfloatsep
\begin{algorithm}[t]
\caption{ID-GNN embedding computation algorithm}
\label{alg:train}
\textbf{Input:} Graph $\mathcal{G}(\mathcal{V}; \mathcal{E})$, input node features $\{x_v, \forall v \in \mathcal{V}\}$; Number of layers $K$; trainable functions $\textsc{Msg}^{(k)}_1(\cdot)$ for nodes with identity coloring, $\textsc{Msg}^{(k)}_0(\cdot)$ for the rest of nodes; $\textsc{Ego}(v, k)$ extracts the $K$-hop ego network centered at node $v$, indicator function $\mathbbm{1}[s = v] = 1 \text{ if }s = v \text{ else } 0$\\
\textbf{Output:} Node embeddings $\mb{h}_v$ for all $v \in \mathcal{V}$

\begin{algorithmic}[1]
\FOR{$v \in \mathcal{V}$}
\STATE $\mathcal{G}_v^{(K)} \leftarrow \textsc{Ego}(v, K)$,\hspace{3mm} $\mb{h}_u^{(0)} \leftarrow \mb{x}_u, \forall u \in \mathcal{G}_v^{(K)}$
\FOR{$k=1,\dots,K$}
    \FOR{$u \in \mathcal{G}_v^{(K)}$}
        \STATE
        $\mb{h}_u^{(k)} \leftarrow \textsc{Agg}^{(k)}\big($ \\ \hspace{4mm}$\{\textsc{Msg}^{(k)}_{\mathbbm{1}[s = v]}(\mb{h}_s^{(k-1)}), s \in \mathcal{N}(u)\}, \mb{h}_u^{(k-1)}\big)$
    \ENDFOR
\ENDFOR
\STATE $\mb{h}_v \leftarrow \mb{h}_v^{(K)}$  
\ENDFOR
\end{algorithmic}
\end{algorithm}

\xhdr{Algorithmic complexity}
Besides adding the identity coloring and applying two types of message passing instead of one, the computation of ID-GNN is almost identical to the widely used mini-batch version of GNNs \cite{hamilton2017inductive,ying2018graph}. In our experiments, by matching the number of trainable parameters, the computation FLOPS used by ID-GNNs and mini-batch GNNs can be the same (shown in Table \ref{tab:run_time}). 

\xhdr{Extension to edge-level tasks}
Here we discuss how to extend the ID-GNN framework to properly resolve existing GNN limitations in edge-level tasks (Figure \ref{fig:overview}, middle). Suppose we want to predict the edge-level label for a node pair $u$, $v$. For ID-GNN, the prediction is made from a \emph{conditional node embedding} $\mb{h}_{u|v}$, which is computed by assigning node $v$, rather than $u$, identity coloring in node $u$'s computation graph, as illustrated in Figure \ref{fig:overview}.
In the case where node $v$ does not lie within $u$'s $K$-hop ego network, no identity coloring is used and ID-GNNs will still suffer from existing failure cases of GNNs. Therefore, we use deeper ID-GNNs for edge-level prediction tasks in practice.

\subsection{ID-GNNs Expressive Power: Theoretical Results}

\xhdr{ID-GNNs are strictly more expressive than existing message passing GNNs}
It has been shown that existing message passing GNNs have an expressive power upper bound by the 1-WL test, where the upper bound can be instantiated by the Graph Isomophism Network (GIN) \cite{xu2018powerful}. 
\begin{proposition}
ID-GNN version of GIN can differentiate any graph that GIN can differentiate, while being able to differentiate certain graphs that GIN fails to distinguish.
\end{proposition}
By setting $\textsc{Msg}^{(k)}_0(\cdot)=\textsc{Msg}^{(k)}_1(\cdot)$, Eq. \ref{eq:idgnn} becomes identical to Eq. \ref{eq:gnn} which trivially proves the first part. The $d$-regular graph example given in Figure \ref{fig:overview} then proves the second part.

\xhdr{ID-GNNs can count cycles}
Proposition 1 provides an overview of the added expressive power of ID-GNNs. Here, we reveal one concrete aspect of this added expressive power, \ie, ID-GNN's capability to count cycles.
We observe that the ability of counting cycles is intuitive to understand; moreover, it is crucial for useful tasks such as predicting node clustering coefficient, which we elaborate in the next section.
\begin{proposition}
For any node $v$, there exists a $K$-layer ID-GNN instantiation that can learn an embedding $\mb{h}_v^{(K)}$ where the $j$-th dimension $\mb{h}_v^{(K)}[j]$ equals the number of length $j$ cycles starting and ending at node $v$, for $j=1,...,K$.
\end{proposition}
We prove this by showing that ID-GNNs can count paths from any node u to the identity node v. Through induction, we show that a 1-layer ID-GNN embedding $\mb{h}_u^{(1)}$ can count length 1 paths from u to v. Then, given a $K$-layer ID-GNN embedding $\mb{h}_u^{(K)}$ that counts paths of length $1,\dots,K$ between u and v, we show the $K+1$-th layer of ID-GNN can accurately update $\mb{h}_v^{(K+1)}$ to account for paths of length $K+1$. Detailed proofs are provided in the Appendix.

\subsection{ID-GNNs Expressive Power: Case Studies}
\label{sec:case_study}

\xhdr{Node-level: Predicting clustering coefficient}
Here we show that existing message passing GNNs fail to inductively predict clustering coefficients purely from graph structure, while ID-GNNs can.
Clustering coefficient is a widely used metric that characterizes the proportion of closed triangles in a node's 1-hop neighborhood \cite{watts1998collective}. 
The node classification failure case in Figure \ref{fig:overview} demonstrates GNNs' inability to predict clustering coefficients, as GNNs fail to differentiate nodes $v_1$ and $v_2$ with clustering coefficient 1 and 0 respectively.
By using one-hot node features, GNNs can overcome this failure mode \cite{hamilton2017inductive}. However, in this case GNNs are \emph{memorizing} the clustering coefficients for each node, since one-hot encodings prevent generalization to unseen graphs.

Based on Proposition 2, ID-GNNs can learn node embeddings $\mb{h}_v^{(K)}$, where $\mb{h}_v^{(K)}[j]$ equals the number of length $j$ cycles starting and ending at node $v$. Given these cycle counts, we can then calculate clustering coefficient $c_v$ of node $v$:
\begin{equation}
    \begin{aligned}
    & c_v = \frac{|\{e_{su}: s,u \in \mathcal{N}(v), e_{su} \in \mathcal{E}\}|}{(d_v)(d_v - 1)/2} \\
    & = \frac{\mb{h}_v^{(K)}[3]}{\mb{h}_v^{(K)}[2] * (\mb{h}_v^{(K)}[2] - 1)}
    \end{aligned}
\end{equation}
where $d_v$ is the degree of node $v$. Since $c_v$ is a continuous function of $\mb{h}_v^{(K)}$, we can approximate it to an arbitrary $\epsilon$ precision with an MLP due to the universal approximation theorem \cite{hornik1989multilayer}.

\xhdr{Edge-level: Predicting reachability or shortest path distance}
Vanilla GNNs make edge-level predictions from pairs of node embeddings \cite{hamilton2017inductive}. However, this type of approaches fail to
predict reachability or shortest path distance (SPD) between node pairs.
For example, two nodes can have the same GNN node embedding, independent of whether they are located in the same connected component.
Although \cite{velikovi2020neural} shows that proper node feature initialization allows for the prediction of reachability and SPD, ID-GNNs present a general solution to this limitation through the use of conditional node embeddings. 
As discussed in ``Extension to edge-level tasks'', we re-formulate edge-level prediction as conditional node-level prediction; consequently, a $K$-layer ID-GNN can predict if node $u\in\mathcal{G}$ is reachable from $v\in\mathcal{G}$ within $K$ hops by using the conditional node embedding $\mb{h}_{u|v}^{(K)}$ via:
\begin{equation}
    \begin{aligned}
    \label{eq:idgnn_reach}
    & \mb{m}_{s|v}^{(k)} = \begin{cases}
    1 & \text{if }\mathbbm{1}[s = v] = 1\\
    \mb{h}_{s|v}^{(k-1)} &  \text{else}
    \end{cases}, \\
    & \mb{h}_{u|v}^{(k)} = \textsc{Max}\big(\{\mb{m}_{s|v}^{(k)}, s \in \mathcal{N}(u)\}\big)
    \end{aligned}
\end{equation}
where $\mb{h}_{u|v}^{(0)}=0, \forall u \in \mathcal{G}$,
and the output $\mb{h}_{u|v}^{(K)}=1$ if an ID-GNN predicts $u$ are reachable from $v$.

\xhdr{Graph-level: Differentiating random $d$-regular graphs}
As is illustrated in Figure \ref{fig:overview}, existing message passing GNNs cannot differentiate random $d$-regular graphs purely from graph structure, as the computation graphs for each node are identical, regardless of the number of layers. 
Here, we show that ID-GNNs can differentiate a significant proportion of random $d$-regular graphs. 
Specifically, we generate 100 non-isomorphic random $d$-regular graphs and consider 3 settings with different graph sizes ($n$) and node degree ($d$). We use up to length $K$ cycle counts, which a $K$-layer ID-GNN can successfully represent (shown in Proposition 2), to calculate the percentage of these $d$-regular graphs that can be differentiated.
Results in Table \ref{tab:rand_graphs} confirm that the addition of identity information can greatly help differentiate $d$-regular graphs.

\begin{table}[t]
\setlength\tabcolsep{3pt}
\centering
\begin{footnotesize}
\caption{Percentage of random $d$-regular graphs that ID-GNNs can differentiate (unique graph representations / total graphs). Note, none of the graphs can be differentiated by 1-WL test or GNNs regardless of the number of layers.}
\label{tab:rand_graphs}
\resizebox{0.9\columnwidth}{!}{
\begin{tabular}{ccccc}
\toprule
& \multicolumn{4}{c}{ID-GNNs} \\ \cmidrule(lr){2-5}
& Layer=3 & Layer=4 & Layer=5 & Layer=6 \\ \midrule
$n$=64, $d$=4, 100 graphs & 11\% & 64\% & 94\%  & 100\% \\
$n$=40, $d$=5, 100 graphs & 14\% & 82\% & 100\% & 100\% \\
$n$=96, $d$=6, 100 graphs   & 21\% & 88\% & 100\% & 100\% \\ \bottomrule
\end{tabular}
}
\end{footnotesize}
\end{table}

\subsection{ID-GNN-Fast: Injecting Identity via Augmented Node Features}
Given that: \textbf{(1)} mini-batch implementations of GNNs have computational overhead when extracting ego networks, which is required by ID-GNNs with heterogeneous message passing, and \textbf{(2)} cycle count information explains an important aspect of the added expressive power of ID-GNNs over existing GNNs, we propose ID-GNN-Fast, where we inject identity information by using cycle counts as augmented node features. Similar cycle count information is also shown to be useful in the context of graph kernels \cite{zhang2018retgk}.
Following the definition in Proposition 3, we use the count of cycles with length $1,\dots,K$ starting and ending at the node $v$ as augmented node feature $\mb{x}_v^+\in \mathbb{R}^K$. These additional features $\mb{x}_v^+$ can be computed efficiently with sparse matrix multiplication via $\mb{x}_v^+[k] = \text{Diag}(A^k)[v]$, where $A$ is the adjacency matrix. We then update the input node attributes for all nodes by concatenating this augmented feature $\mb{x}_v = \textsc{Concat}(\mb{x}_v, \mb{x}_v^+)$.

\section{Experiments}

\begin{table*}[t]
\setlength\tabcolsep{3pt}
\centering
\begin{footnotesize}
\caption{\textbf{Comparing ID-GNNs with GNNs on graph property prediction tasks}. For each column, all the 12 models have \emph{the same computational budget}. The best performance in each family of models is bold. Results are averaged over 3 random splits.}
\label{tab:graph_property}
\resizebox{\textwidth}{!}{
\begin{tabular}{cccccccccccc}
\toprule

& & \multicolumn{4}{c}{\shortstack{Node classification:\\ predict clustering coefficient}} & \multicolumn{4}{c}{\shortstack{Edge classification:\\ predict shortest path distance}} & \multicolumn{2}{c}{\shortstack{Graph classification:\\predict clustering coefficient}} \\  \cmidrule(lr){3-6}\cmidrule(lr){7-10} \cmidrule(lr){11-12} 
& & \texttt{ScaleFree} & \texttt{SmallWorld} & \texttt{ENZYMES} & \texttt{PROTEINS} & \texttt{ScaleFree} & \texttt{SmallWorld} & \texttt{ENZYMES} & \texttt{PROTEINS} &  \texttt{ScaleFree} & \texttt{SmallWorld} \\ \midrule

\multirow{4}{*}{GNNs} & GCN & 0.679$\pm$0.01 & \textbf{0.589$\pm$0.04} & 0.596$\pm$0.02 & 0.540$\pm$0.00 & 0.522$\pm$0.00 & 0.558$\pm$0.02 & \textbf{0.557$\pm$0.02} & \textbf{0.722$\pm$0.01} & 0.270$\pm$0.06 & 0.433$\pm$0.03 \\ 
& SAGE & 0.470$\pm$0.03 & 0.271$\pm$0.03 & 0.572$\pm$0.04 & 0.444$\pm$0.03 & 0.297$\pm$0.01 & 0.360$\pm$0.16 & 0.550$\pm$0.02 & 0.722$\pm$0.01 & 0.047$\pm$0.03 & 0.077$\pm$0.01 \\ 
& GAT & 0.470$\pm$0.03 & 0.274$\pm$0.06 & 0.464$\pm$0.03 & 0.400$\pm$0.02 & 0.451$\pm$0.00 & 0.551$\pm$0.02 & 0.556$\pm$0.02 & 0.722$\pm$0.01 & 0.127$\pm$0.04 & 0.093$\pm$0.03 \\ 
& GIN & \textbf{0.693$\pm$0.00} & 0.571$\pm$0.04 & \textbf{0.660$\pm$0.02} & \textbf{0.558$\pm$0.02} & \textbf{0.551$\pm$0.01} & \textbf{0.575$\pm$0.02} & 0.541$\pm$0.03 & 0.722$\pm$0.01 & \textbf{0.280$\pm$0.01} & \textbf{0.453$\pm$0.03} \\ \midrule 
\multirow{4}{*}{\textbf{\shortstack{ID-GNNs\\Fast}}} & GCN & 0.897$\pm$0.01 & 0.812$\pm$0.02 & 0.786$\pm$0.04 & 0.805$\pm$0.02 & 0.521$\pm$0.00 & 0.576$\pm$0.02 & 0.553$\pm$0.03 & 0.722$\pm$0.01 & 0.823$\pm$0.04 & \textbf{0.850$\pm$0.06} \\ 
& SAGE & \textbf{0.954$\pm$0.01} & \textbf{0.994$\pm$0.00} & \textbf{0.958$\pm$0.04} & \textbf{0.985$\pm$0.01} & 0.527$\pm$0.01 & \textbf{0.583$\pm$0.02} & 0.551$\pm$0.04 & \textbf{0.722$\pm$0.01} & \textbf{0.827$\pm$0.02} & 0.810$\pm$0.04 \\ 
& GAT & 0.889$\pm$0.01 & 0.739$\pm$0.03 & 0.675$\pm$0.04 & 0.675$\pm$0.05 & 0.471$\pm$0.00 & 0.574$\pm$0.02 & 0.545$\pm$0.03 & 0.722$\pm$0.01 & 0.620$\pm$0.01 & 0.800$\pm$0.06 \\ 
& GIN & 0.895$\pm$0.00 & 0.822$\pm$0.03 & 0.798$\pm$0.06 & 0.790$\pm$0.01 & \textbf{0.546$\pm$0.00} & 0.576$\pm$0.02 & \textbf{0.556$\pm$0.03} & 0.722$\pm$0.01 & 0.730$\pm$0.02 & 0.840$\pm$0.06 \\ \midrule 
\multirow{4}{*}{\textbf{\shortstack{ID-GNNs\\Full}}} & GCN & \textbf{\emph{0.985$\pm$0.01}} & \textbf{\emph{0.994$\pm$0.00}} & 0.984$\pm$0.02 & \textbf{\emph{0.995$\pm$0.00}} & 0.999$\pm$0.00 & 1.000$\pm$0.00 & 0.994$\pm$0.00 & 0.998$\pm$0.00 & \textbf{\emph{0.830$\pm$0.03}} & \textbf{\emph{0.877$\pm$0.05}} \\ 
& SAGE & 0.588$\pm$0.01 & 0.400$\pm$0.12 & 0.591$\pm$0.04 & 0.474$\pm$0.03 & \textbf{\emph{1.000$\pm$0.00}} & \textbf{\emph{1.000$\pm$0.00}} & \textbf{\emph{1.000$\pm$0.00}} & 1.000$\pm$0.00 & 0.247$\pm$0.01 & 0.250$\pm$0.10 \\ 
& GAT & 0.638$\pm$0.01 & 0.847$\pm$0.20 & \textbf{\emph{0.994$\pm$0.00}} & 0.994$\pm$0.00 & 0.984$\pm$0.00 & 0.989$\pm$0.01 & 0.963$\pm$0.01 & 0.993$\pm$0.01 & 0.047$\pm$0.03 & 0.067$\pm$0.01 \\ 
& GIN & 0.716$\pm$0.01 & 0.572$\pm$0.04 & 0.655$\pm$0.02 & 0.570$\pm$0.03 & 1.000$\pm$0.00 & 0.964$\pm$0.05 & 1.000$\pm$0.00 & \textbf{\emph{1.000$\pm$0.00}} & 0.273$\pm$0.02 & 0.490$\pm$0.01 \\ \midrule 
\multicolumn{2}{c}{\textbf{Best ID-GNN over best GNN}}& 29.3\%& 40.6\%& 33.4\%& 43.7\%& 44.9\%& 42.5\%& 44.3\%& 27.8\%& 55.0\%& 42.3\% \\

\bottomrule
\end{tabular}
}
\end{footnotesize}
\end{table*}

\begin{table*}[t]
\setlength\tabcolsep{3pt}
\centering
\begin{footnotesize}
\caption{\textbf{Comparing GNNs with ID-GNNs on real-world prediction tasks}. For each column, all the 12 models have \emph{the same computational budget}. The best performance in each family of models is bold. Results are averaged over 3 random splits.}
\label{tab:real_world}
\resizebox{\textwidth}{!}{
\begin{tabular}{cccccccccccc}
\toprule

& & \multicolumn{2}{c}{\shortstack{Node classification:\\ real-world labels}} & \multicolumn{4}{c}{\shortstack{Edge classification:\\ link prediction}} & \multicolumn{4}{c}{\shortstack{Graph classification:\\ real-world labels}} \\  \cmidrule(lr){3-4}\cmidrule(lr){5-8} \cmidrule(lr){9-12} 
& & \texttt{Cora} & \texttt{CiteSeer} & \texttt{ScaleFree} & \texttt{SmallWorld} & \texttt{ENZYMES} & \texttt{PROTEINS} & \texttt{ENZYMES} & \texttt{PROTEINS} &  \texttt{BZR} & \texttt{ogbg-molhiv} \\ \midrule

\multirow{4}{*}{GNNs} & GCN & 0.848$\pm$0.01 & 0.709$\pm$0.01 & 0.796$\pm$0.01 & 0.709$\pm$0.00 & 0.651$\pm$0.01 & 0.659$\pm$0.01 & 0.547$\pm$0.01 & 0.695$\pm$0.02 & 0.844$\pm$0.04 & 0.747$\pm$0.02 \\ 
& SAGE & \textbf{0.868$\pm$0.01} & \textbf{0.726$\pm$0.01} & 0.541$\pm$0.00 & 0.512$\pm$0.00 & 0.546$\pm$0.01 & 0.582$\pm$0.01 & 0.542$\pm$0.01 & 0.692$\pm$0.01 & 0.852$\pm$0.04 & 0.758$\pm$0.01 \\ 
& GAT & 0.857$\pm$0.01 & 0.716$\pm$0.01 & 0.500$\pm$0.00 & 0.500$\pm$0.00 & 0.478$\pm$0.01 & 0.491$\pm$0.01 & \textbf{0.555$\pm$0.02} & \textbf{0.723$\pm$0.00} & 0.848$\pm$0.03 & 0.742$\pm$0.01 \\ 
& GIN & 0.858$\pm$0.01 & 0.719$\pm$0.01 & \textbf{0.802$\pm$0.01} & \textbf{0.722$\pm$0.01} & \textbf{0.654$\pm$0.01} & \textbf{0.667$\pm$0.00} & 0.553$\pm$0.02 & 0.721$\pm$0.00 & \textbf{0.856$\pm$0.02} & \textbf{0.762$\pm$0.03} \\ \midrule 
\multirow{4}{*}{\textbf{\shortstack{ID-GNNs\\Fast}}} & GCN & 0.851$\pm$0.02 & 0.715$\pm$0.00 & 0.856$\pm$0.03 & 0.719$\pm$0.00 & 0.649$\pm$0.01 & 0.671$\pm$0.01 & 0.600$\pm$0.01 & \textbf{\emph{0.741$\pm$0.02}} & 0.807$\pm$0.02 & 0.772$\pm$0.02 \\ 
& SAGE & 0.866$\pm$0.02 & \textbf{\emph{0.742$\pm$0.01}} & \textbf{\emph{0.898$\pm$0.01}} & 0.743$\pm$0.02 & 0.671$\pm$0.04 & 0.701$\pm$0.01 & \textbf{\emph{0.639$\pm$0.00}} & 0.724$\pm$0.03 & 0.835$\pm$0.06 & \textbf{0.780$\pm$0.01} \\ 
& GAT & \textbf{0.870$\pm$0.02} & 0.719$\pm$0.02 & 0.731$\pm$0.02 & 0.537$\pm$0.00 & 0.490$\pm$0.01 & 0.502$\pm$0.01 & 0.619$\pm$0.03 & 0.715$\pm$0.03 & 0.848$\pm$0.05 & 0.740$\pm$0.01 \\ 
& GIN & 0.864$\pm$0.01 & 0.719$\pm$0.01 & 0.837$\pm$0.01 & \textbf{0.759$\pm$0.01} & \textbf{0.718$\pm$0.02} & \textbf{0.724$\pm$0.00} & 0.567$\pm$0.01 & 0.723$\pm$0.01 & \textbf{0.864$\pm$0.03} & 0.755$\pm$0.02 \\ \midrule 
\multirow{4}{*}{\textbf{\shortstack{ID-GNNs\\Full}}} & GCN & 0.863$\pm$0.01 & 0.719$\pm$0.01 & 0.771$\pm$0.04 & 0.798$\pm$0.03 & 0.838$\pm$0.01 & 0.878$\pm$0.02 & \textbf{0.586$\pm$0.04} & 0.715$\pm$0.02 & 0.881$\pm$0.04 & 0.769$\pm$0.01 \\ 
& SAGE & 0.875$\pm$0.01 & \textbf{0.730$\pm$0.02} & 0.741$\pm$0.01 & 0.724$\pm$0.03 & 0.819$\pm$0.01 & 0.863$\pm$0.01 & 0.547$\pm$0.02 & 0.721$\pm$0.01 & 0.864$\pm$0.02 & \textbf{\emph{0.783$\pm$0.02}} \\ 
& GAT & \textbf{\emph{0.878$\pm$0.01}} & 0.729$\pm$0.01 & 0.749$\pm$0.01 & 0.742$\pm$0.03 & 0.824$\pm$0.01 & 0.859$\pm$0.03 & 0.567$\pm$0.05 & \textbf{0.738$\pm$0.01} & \textbf{\emph{0.881$\pm$0.04}} & 0.739$\pm$0.01 \\ 
& GIN & 0.851$\pm$0.00 & 0.725$\pm$0.01 & \textbf{0.815$\pm$0.01} & \textbf{\emph{0.810$\pm$0.03}} & \textbf{\emph{0.846$\pm$0.01}} & \textbf{\emph{0.886$\pm$0.02}} & 0.544$\pm$0.02 & 0.730$\pm$0.03 & 0.852$\pm$0.03 & 0.756$\pm$0.00 \\ \midrule 
\multicolumn{2}{c}{\textbf{Best ID-GNN over best GNN}}& 1.0\%& 1.6\%& 9.6\%& 8.7\%& 19.2\%& 21.9\%& 8.3\%& 1.8\%& 2.5\%& 2.0\% \\

\bottomrule
\end{tabular}
}
\end{footnotesize}
\end{table*}

\subsection{Experimental setup}

\xhdr{Datasets}
We perform experiments over 8 different datasets.
We consider the synthetic graph datasets \textbf{(1)} \texttt{ScaleFree} \cite{holme2002growing} and \textbf{(2)} \texttt{SmallWorld} \cite{watts1998collective}, each containing 256 graphs, with average degree of $4$ and average clustering coefficient in the range $[0,0.5]$.
For real-world datasets we explore 3 protein datasets: \textbf{(3)} \texttt{ENZYMES} \cite{borgwardt2005protein} with 600 graphs, \texttt{(4)} \texttt{PROTEINS} \cite{schomburg2004brenda} with 1113 graphs, and \textbf{(5)} \texttt{BZR} \cite{sutherland2003spline} with 405 graphs. We also consider citation networks including \textbf{(6)} \texttt{Cora} and \textbf{(7)} \texttt{CiteSeer} \cite{sen2008collective}, and a large-scale molecule dataset \textbf{(8)} \texttt{ogbg-molhiv} \cite{hu2020open} with 41K graphs. 

\xhdr{Tasks}
We evaluate ID-GNNs over two task categories. First, we consider challenging graph property prediction tasks:
\textbf{(1)} classifying nodes by clustering coefficients, \textbf{(2)} classifying pairs of nodes by their shortest path distances, and \textbf{(3)} classifying random graphs by their average clustering coefficients. We bin over continuous clustering coefficients to make task (1) and (3) 10-way classification tasks and threshold the shortest path distance to make task (2) a 5-way classification task.
We also consider more common tasks with real-world labels, including \textbf{(4)} node classification, \textbf{(5)} link prediction, and \textbf{(6)} graph classification.
For the \texttt{ogbg-molhiv} dataset we use provided splits, while for all the other tasks, we use a random 80/20\% train/val split and average results over 3 random splits. 
Validation accuracy (multi-way classification) or ROC AUC (binary classification) in the final epoch is reported.

\xhdr{Models}
We present a standardized framework for fairly comparing ID-GNNs with existing GNNs. 
We use 4 widely adopted GNN models as base models: GAT \cite{velickovic2017graph}, GCN \cite{kipf2016semi}, GIN \cite{xu2018powerful}, and GraphSAGE \cite{hamilton2017inductive}.
We then transform each GNN model to its ID-GNN variants, ID-GNN-Full (based on heterogeneous message passing) and ID-GNN-Fast, holding all the other hyperparameters fixed. To further ensure fairness, we adjust layer widths, so that all the models match the number of trainable parameters of a standard GCN model (i.e., match computational budget).
In summary, we run 12 models for each experimental setup, including 4 types of GNN architectures, each with 3 versions.

We use 3-layer GNNs for node and graph level tasks, and 5-layer GNNs for edge level tasks, where GCNs with 256-dim hidden units are used to set the computational budget for all 12 model variants. For ID-GNNs-Full, each layer has 2 sets of weights, thus each layer has fewer number of hidden units; for ID-GNNs-Fast, 10-dim augmented cycle counts features are used.
We use ReLU activation and Batch Normalization for all the models.
We use Adam optimizer with learning rate 0.01. Due to the different nature of these tasks, tasks (1)(3)(6) excluding the \texttt{ogbg-molhiv} dataset, are trained for 1000 epochs, while the rest are trained for 100 epochs.
For node-level tasks, GNN / ID-GNN node embeddings are directly used for prediction; for edge-level tasks, ID-GNNs-Full make predictions with conditional node embeddings, while GNNs and ID-GNNs-Fast make predictions by concatenating pairs of node embeddings and then passing the result through a 256-dim MLP; for graph-level tasks, predictions are based on a global sum pooling over node embeddings. 

Overall, these comprehensive and consistent experimental settings reveal the general improvement of ID-GNNs compared with existing GNNs.

\subsection{Graph Property Prediction Tasks}

\xhdr{Node clustering coefficient prediction}
In Table \ref{tab:graph_property} we observe that across all models and datasets, both ID-GNN formulations perform at the level of or significantly outperform GNN counterparts, with an average absolute performance gain of 36.8\% between the best ID-GNN and best GNN. In each dataset, both ID-GNN methods perform with near 100\% accuracy for at least one GNN architecture.
ID-GNN-Fast shows the most consistent improvements across models with greatest improvement in GraphSAGE. These results align with the previous discussion of using cycle counts alone to learn clustering coefficients. We defer discussion until later on ID-GNN-Full sometimes showing minimal improvement, to present a general understanding of this behavior. 

\xhdr{Shortest path distance prediction}
In the pairwise shortest path prediction task, ID-GNNs-Full outperform GNNs by an average of 39.9\%. Table \ref{tab:graph_property} reveals that ID-GNN-Full performs with 100\% or near 100\% accuracy under all GNN architectures, across all datasets. This observation, along with the comparatively poor performance of ID-GNNs-Fast and GNNs, confirms the previously discussed conclusion that traditional edge-level predictions, through pairwise node embeddings, fail to accurately make edge-level predictions.

\xhdr{Average clustering coefficient prediction for random graphs}
In Table \ref{tab:graph_property}, we observe that adding identity information results in a 55\% and 42.3\% increase in best performance over \texttt{ScaleFree} and \texttt{SmallWorld} graphs respectively. ID-GNN-Fast shows the most consistent improvement (56.9\% avg. model gain),
which aligns with previous intuitions about the utility of cycle count information in predicting clustering coefficients and differentiating random graphs.

\subsection{Real-world Prediction Tasks}
\xhdr{Node classification}
In node classification we see smaller but still significant improvements when using ID-GNNs. Table \ref{tab:real_world} shows an overall 1\% and 1.6\% improvement for \texttt{Cora} and \texttt{CiteSeer} respectively. In all cases except for GIN and GraphSAGE on \texttt{Cora}, adding identity information improves performance. 
In regards to the relatively small improvements, we hypothesize that the richness of node features (over 1000-dim for both datasets) greatly dilutes the importance of graph structure in these tasks, and thus the added expressiveness from identity information is diminished.

\xhdr{Link prediction}
As shown in Table \ref{tab:real_world}, we observe consistent improvement in ID-GNNs over GNNs, with 9.2\% and 20.6\% ROC AUC improvement on synthetic and real-world graphs respectively. Moreover, we observe that ID-GNN-Full nearly always performs the best, aligning with previous edge-level task results in Table \ref{tab:graph_property} and intuitions on the importance of re-formulating edge-level tasks as conditional node prediction tasks. We observe that performance improves less for random graphs, which we hypothesize is due to the randomness within these synthetic graphs causing the distinction between positive and negative edges to be much more vague.

\xhdr{Graph classification}
Across each dataset, we observe that the best ID-GNN consistently outperforms the best GNN of the same computational budget.
However, model to model improvement is less clear. For the \texttt{ENZYMES} dataset, ID-GNN-Fast shows strong improvements under each GNN architecture, with gains as large as 10\% in accuracy for the GraphSAGE model. In \texttt{PROTEIN} and \texttt{BZR}, ID-GNN-Full shows improvements for each GNN model (except GIN on \texttt{BZR}), with greatest performance increases in GCN and GAT (avg. 3.6\% and 3.0\% respectively).

\begin{table}[!tbp]
\setlength\tabcolsep{2pt}
\centering
\begin{footnotesize}
\caption{Runtime analysis for GCN and ID-GNN equivalents given the same computational budget. For each model, average time (milisecond) per batch of 128 \texttt{ENZYME} graphs is reported for the forward and the forward + backward pass.}
\label{tab:run_time}
\resizebox{\columnwidth}{!}{
\begin{tabular}{ccccc}
\toprule
& GCN & ID-GNN-Fast & GCN (mini-batch) & ID-GNN-Full \\ \midrule
forward & 4.8$\pm$0.1 & 4.9$\pm$0.1 & 28.1$\pm$0.1 & 24.2$\pm$4.0 \\
forward + backward & 8.9$\pm$0.7 &  10.0$\pm$0.6 & 33.3$\pm$0.9 & 31.1$\pm$0.8 \\ \bottomrule
\end{tabular}
}
\end{footnotesize}
\end{table}

\subsection{Computational Cost Analysis}
We compare the runtime complexity (excluding mini-batch loading time) of ID-GNNs vs. existing GNNs, where we hold the computational budget constant across all models. Table \ref{tab:run_time} reveals that when considering the forward and backward pass, ID-GNN-Full runs 3.8x slower than its GNN equivalent but has an equivalent runtime complexity to the mini-batch implementation of GNN, while ID-GNN-Fast runs with essentially zero overhead over existing GNN implementations. 

\subsection{Summary of Comparisons with GNNs}
\label{sec:findings}
Overall, ID-GNN-Full and ID-GNN-Fast demonstrate significant improvements over their message passing GNN counterparts, of the same computational budget, on a variety of tasks.
In all tasks, the best ID-GNNs outperforms the best GNNs; moreover, out of 160 model-task combinations, ID-GNNs fail to improve accuracy in fewer than 10 cases. For the rare cases where there is no improvement from ID-GNN-Full, we suspect that the model \emph{underfits} since we control the complexity of models: given that ID-GNN-Full has two sets of weights (heterogeneous message passing), fewer weights are used for each message passing. 
For verification, if we double the computational budget, we observe that ID-GNN versions again outperform GNN counterparts.

\subsection{Comparisons with Expressive Graph Networks}
We provide additional experimental comparisons against other expressive graph networks in both edge and graph-level tasks. For edge-level task, we further compare with P-GNN \cite{you2019position} over the \texttt{ENZYMES} and \texttt{PROTEINS} datasets using the protocol introduced previously. For graph-level comparison, we include experimental results over 3 datasets: \texttt{MUTAG} with 182 graphs \cite{debnath1991structure}, \texttt{PTC} with 344 graphs \cite{helma2003survey}, and \texttt{PROTEINS}. We follow PPGN's \cite{maron2019provably} 10-fold 90/10 data splits and compare against 5 other expressive graph networks. We report numbers in the corresponding papers, and report the best ID-GNNs out of the 4 variants.

\xhdr{Link prediction}
We compare against P-GNNs on 2 link prediction datasets. As shown in the Table \ref{tab:More_Powerful_Link}, we observe significant improvements using ID-GNNs compared to both its GNN counterpart and P-GNNs. 
These results both demonstrate ID-GNNs' competitive performance as a general graph learning method against a task-specific model, while also highlighting ID-GNN's improved ability to incorporate node-features compared with P-GNNs.  

\xhdr{Graph classification}
We compare ID-GNNs against several other more powerful graph networks in the task of graph classification. Table \ref{tab:More_Powerful_Graph} demonstrates the strong performance of ID-GNNs. ID-GNNs outperform other graph networks on the \texttt{MUTAG} and \texttt{PROTEINS} datasets; Although ID-GNNs performance then drops on the \texttt{PTC} dataset, they are still comparable to two out of the four powerful graph models. These strong results further demonstrate the ability of ID-GNN to outperform not only message passing GNNs, but also other powerful, task specific graph networks across a range of tasks.

\begin{table}[!tbp]
\setlength\tabcolsep{3pt}
\centering
\begin{footnotesize}
\caption{Comparisons with P-GNN on link prediction task.}
\vspace{-1mm}
\label{tab:More_Powerful_Link}
\resizebox{\columnwidth}{!}{
\begin{tabular}{ccccc}
\toprule
& \multicolumn{2}{c}{Edge classification: link prediction} \\ \cmidrule(lr){2-3}
 & \texttt{ENZYMES} & \texttt{PROTEINS} \\ \midrule
Best GNN & 0.654$\pm$0.015 & 0.667$\pm$0.002 \\
P-GNN \cite{you2019position} & 0.715$\pm$0.024 &  0.810$\pm$0.013 \\ \midrule
Best ID-GNN-Fast & 0.718$\pm$0.010 & 0.724$\pm$0.015 \\
Best ID-GNN-Full & \textbf{0.846$\pm$0.010} & \textbf{0.886$\pm$0.015} \\
\bottomrule
\end{tabular}
}
\end{footnotesize}
\end{table}

\begin{table}[!tbp]
\setlength\tabcolsep{2pt}
\centering
\begin{footnotesize}
\caption{Comparisons with other expressive graph networks on graph classification tasks. We use evaluation setup from \cite{maron2019provably}, and the reported numbers in corresponding papers are shown.}
\vspace{-1mm}
\label{tab:More_Powerful_Graph}
\resizebox{\columnwidth}{!}{
\begin{tabular}{ccccc}
\toprule
& \multicolumn{3}{c}{Graph classification: real-world labels} \\ \cmidrule(lr){2-4}
 & \texttt{MUTAG} & \texttt{PTC} & \texttt{PROTEINS} \\ \midrule
Best GNN & 0.905$\pm$0.057 & 0.617$\pm$0.046 & 0.773$\pm$0.037 \\ 
PPGN \cite{maron2019provably} & 0.906$\pm$0.087 & 0.662$\pm$0.065 & 0.772$\pm$0.047 \\ 
CCN \cite{kondor2018covariant} & 0.916$\pm$0.072 & \textbf{0.706 $\pm$0.07} & NA\\
1-2-3 GNN \cite{morris2019weisfeiler} & 0.861 & 0.609 & 0.755  \\
Invariant GNs \cite{maron2018invariant} & 0.846$\pm$0.10 & 0.595 $\pm$ 0.073 & 0.752 $\pm$ 0.043\\
GSN \cite{bouritsas2020improving} & 0.922 $\pm$ 0.075 & 0.682 $\pm$ 0.072 & 0.766 $\pm$ 0.050\\
\midrule
Best ID-GNN-Fast & \textbf{0.965$\pm$0.032} & 0.619$\pm$0.054 & \textbf{0.780$\pm$0.035} \\
Best ID-GNN-Full & \textbf{0.930$\pm$0.056} & 0.625$\pm$0.053 & \textbf{0.779$\pm$0.024} \\
\bottomrule
\end{tabular}
}
\end{footnotesize}
\end{table}

\section{Conclusion}
We have proposed ID-GNNs as a general and powerful extension to existing GNNs with rich theoretical and experimental results.
Specifically, ID-GNNs have expressive power beyond the 1-WL test. When runtime efficiency is the primary concern, we also present a feature augmented version of ID-GNN that maintains theoretical guarantees and empirical success of heterogeneous message passing, while only requiring one-time feature pre-processing. We recommend that this cycle-count feature augmentation be the new go-to node feature initialization when additional node attributes are not available. Additionally, as direct extensions to message passing GNNs, ID-GNNs can be easily implemented and extended via existing code platform. Overall, ID-GNNs outperform corresponding message passing GNNs, while both maintaining the attractive proprieties of message passing GNNs and demonstrating competitive performance compared with other powerful/expressive graph networks. We hope ID-GNNs' added expressive power and proven practical applicability can enable exciting new applications and further development of message passing GNNs.

\section*{Ethics Statement}

GNNs represent a promising family of models for analyzing and understanding relational data. A broad range of application domains, such as network fraud detection \cite{akoglu2013opinion, kumar2017antisocial, akoglu2013anomaly}, molecular drug structure discovery \cite{you2018graph,you2018graphrnn,jin2018junction}, recommender systems \cite{ying2018graph,you2019hierarchical}, and network analysis \cite{kumar2017antisocial, morris2019weisfeiler, fan2019graph, ying2019gnnexplainer} stand to be greatly impacted by the use and development of GNNs.
As a direct extension of existing message passing GNNs, ID-GNNs represent a simple but powerful transformation to GNNs that re-frames the discussion on GNN expressive power and thus their performance in impactful problem domains. In comparison to other models that have expressive power beyond 1-WL tests,
ID-GNNs are easy to implement with existing graph learning packages; therefore, ID-GNNs can be easily used as extensions of existing GNN models for tackling important real-world tasks, as well as themselves extended and further explored in the research space. 

The simplicity of ID-GNNs presents great promise for further exploration into the expressiveness of GNNs. In particular, we believe that our work motivates further research into heterogeneous message passing and coloring schemes, as well as generic, but powerful forms of feature augmentation. By further increasing the expressiveness of message passing GNNs, we hopefully enable new, important tasks to be solved across a wide range of disciplines or significant improvement on previously defined and widely adopted GNN models. Through ease of use and strong preliminary results, we believe that our work opens the doors for new explorations into the study of graphs and graph based tasks, with the potential for great improvement in existing GNN models.

\section*{Acknowledgments}
We gratefully acknowledge the support of
DARPA under Nos. FA865018C7880 (ASED), N660011924033 (MCS);
ARO under Nos. W911NF-16-1-0342 (MURI), W911NF-16-1-0171 (DURIP);
NSF under Nos. OAC-1835598 (CINES), OAC-1934578 (HDR), CCF-1918940 (Expeditions), IIS-2030477 (RAPID);
Stanford Data Science Initiative, 
Wu Tsai Neurosciences Institute,
Chan Zuckerberg Biohub,
Amazon, Boeing, JPMorgan Chase, Docomo, Hitachi, JD.com, KDDI, NVIDIA, Dell. 
J. L. is a Chan Zuckerberg Biohub investigator.
Jiaxuan You is supported by JPMorgan Chase PhD Fellowship and Baidu Scholarship.
Rex Ying is supported by Baidu Scholarship.

\bibliography{bibli}

\appendix
\section{Proof of Proposition 2}

Consider an arbitrary node $v$ in a graph $G = (V, E)$ for which we want to compute the embedding under ID-GNN. Without loss of generality assume that $x_u = [1], \forall u \in V$. 
Additionally, let $\textsc{Msg}^{(k)}_0(\cdot)$, $\textsc{Msg}^{(k)}_1(\cdot)$, and $\textsc{Agg}^{(k)}(\cdot)$,
from main paper Eq. 3  (the general formulation for the k-th layer of heterogeneous message passing), be defined as follows:
\begin{equation}
    \begin{aligned}
    & \textsc{Msg}^{(k)}_0(\cdot) = W_0^k h_u^{k-1} + b_0^k, \quad
    \textsc{Msg}^{(k)}_1(\cdot) = W_1^k h_u^{k-1} + b_1^k,\\
    & \textsc{Agg}^{(k)}(\cdot) = \textsc{SUM}(\{m_u^k, u \in N(w)\})
    \end{aligned}
\end{equation}
where $W_0^k, b^k_0$ are trainable weights for nodes without identity coloring, and $W_1^k, b^k_1$ are for nodes with identity coloring. Assume the following weight matrix assignments for different layers k:

\textbf{k = 1:} Let $W_0^1 = [0]$ (i.e the 0 matrix), $b^0_1 = [0]$, $W_1^1 = [0]$, and $b^1_1 = [1]$.

\textbf{k = $\mathbf{2, \dots, K}$:} Let $W_0^k = W_1^k = [0, I]^T \in \mathbb{R}^{k \times (k - 1)}$ with identity matrix $I \in \mathbb{R}^{(k-1) \times (k-1)}$,  $b^k_0 = [0, ..., 0]^T \in \mathbb{R}^k$, and $b^k_1 = [1, 0,...,0]^T \in \mathbb{R}^k$.

We will first prove Lemma 1 by induction.

\textbf{Lemma 1:} \textit{After n-layers of heterogeneous message passing w/r to the identity colored node $v$, the embedding $h_u^n \in R^n, \forall u \in V$ is such that $h_u^n[j] = $ the number of paths of length ($j$) starting at node $u$ and ending at the identity node $v$ for $j = 1,...,n$.}

\textbf{Proof of Lemma 1:}

\textbf{Base case:} We consider the result after 1 layer of message passing. For an arbitrary node u, we see that:
\begin{equation}
    h^1_u = \sum_{w \in N(u)} \big([0] h_w^{k-1} + [1]\mathbbm{1}[w = v]\big) = \sum_{w \in N(u)} [1]\mathbbm{1}[w = v]
\end{equation}

where the indicator function $\mathbbm{1}[w = v] = 1$ if $w=v$ else 0 is used to reflect node coloring. We see that $h^1_u[1] =$ exactly the number of paths of length 1 from u to the identity node v.

\textbf{Inductive Hypothesis:} We assume that after k-layers of heterogeneous message passing the embedding $h_u^k \in R^k, \forall u \in V$ is such that $h_u^k[j] = $ the number of paths of length j starting at node $u$ and ending at the identity node $v$ for $j = 1,...,k$. We will prove that after one more layer of message passing or the $(k+1)$th layer of ID-GNN the desired property still holds for the updated embeddings $h_u^{k+1} \in R^{k+1}, \forall u \in V$. To do so we consider the update for an arbitrary node $u$:
\begin{equation}
    \begin{aligned}
    h_u^{k+1} &= \sum_{w \in N(u)} \big([0, I]^T h_w^{k} + [1, 0,...,0]^T \mathbbm{1}[w = v]\big)  \\&= \sum_{w \in N(u)} [\mathbbm{1}[w = v], h_w^{k}]^T
    \end{aligned}
\end{equation}
We see that $h_u^{k+1}[1] =$ the number of length 1 paths to the identity node v, and for $j = 2,...,k+1 \rightarrow h_u^{k+1}[j] = \sum_{w \in N(u)}h_w^{k}[j - 1]$, which by our inductive hypothesis is exactly the number of paths of length $j$ from node u to v. To see this, we notice that for the node u, we sum up all of the paths of length $j-1$ from the neighboring nodes of u to the destination node v in order to get the number of paths of length j from u to v. Moreover, we see that each path of length $j-1$ is now 1 step longer after an extra layer of message passing giving the desired paths of length $j$. We have thus shown that after $k+1$ layers of message passing the embedding $h_u^{k+1}$ has the desired properties completing induction.

Proposition 2 directly follows from the result of Lemma 1. Namely, if we choose the identity node $v$ itself, by Lemma 1 we can learn an embedding such that $h_v^k$ satisfies $h_v^k[j]=$ the number of length j paths from $v$ to $v$, or equivalently the number of cycles starting and ending at node $v$, for $j=1,\dots,k$.  

\section{Memory consumption of ID-GNN}
Here we discuss the potential memory consumption overhead of ID-GNNs compared with mini-batch GNNs.
In ID-GNN, we adopted GraphSAGE \cite{hamilton2017inductive} style of mini-batch GNN implementation, where disjoint extracted ego-nets are fed into a GNN. Given that we control the computational complexity, ID-GNN-Full has no memory overhead compared with GraphSAGE-style mini-batch GNN. This argument is supported by Table 4 (run-time analysis) in the main manuscript as well.

Mini-batch GNNs can be made more memory efficient by leveraging overlap within the extracted ego-nets; consequently, the embeddings of these nodes only need to be computed once. Note that while this approach saves the embedding computation of GNNs, it increases  the time for processing mini-batches, since nodes from different ego-nets need to be aligned and deduplicated. Comparing the memory usage of ID-GNNs with this memory-efficient mini-batch GNN, the increase in memory usage is moderate. Comparison over \texttt{CiteSeer} reveals that a 3-layer ID-GNN takes 34\%, 79\% and 162\% more memory under batch sizes 16, 32 and 64. Moreover, since mini-batch GNNs are often used for graphs larger than \texttt{CiteSeer}, where the percentage of common nodes between ego-nets is likely small, the overhead of ID-GNN’s memory consumption will be even lower.

\end{document}